\documentclass[twoside,twocolumn,10pt]{article}

\usepackage{wscg}           
\RequirePackage{ifpdf}
\ifpdf
 \RequirePackage[pdftex]{graphicx}
 \RequirePackage[pdftex]{color}
\else
 \RequirePackage[dvips,draft]{graphicx}
 \RequirePackage[dvips]{color}
\fi
\usepackage[switch]{lineno} 

\usepackage{calc}
\title{Why Existing Multimodal Crowd Counting Datasets Can Lead to Unfulfilled Expectations in Real-World Applications}

\author{
\parbox{0.4\textwidth}{\centering
Martin Thi{\ss}en\\[1mm]
Hochschule Darmstadt\\
}
\parbox{0.4\textwidth}{\centering
Prof. Dr. Elke Hergenr{\"o}ther\\[1mm]
Hochschule Darmstadt\\
}
}

\usepackage{url}
\makeatletter
\def\Uslash{\mathbin{\mathchar`\/}\@ifnextchar{/}{\kern-.15em}{}}
\g@addto@macro\UrlSpecials{\do \/ {\Uslash}}
\def\Ucolon{\mathbin{\mathchar`:}\@ifnextchar{/}{\kern-.1em}{}}
\g@addto@macro\UrlSpecials{\do : {\Ucolon}}
\makeatother

\begin{document}

\twocolumn[{\csname @twocolumnfalse\endcsname

\maketitle  

\begin{abstract}
\noindent
More information leads to better decisions and predictions, right? Confirming this hypothesis, several studies concluded that the simultaneous use of optical and thermal images leads to better predictions in crowd counting. However, the way multimodal models extract enriched features from both modalities is not yet fully understood. Since the use of multimodal data usually increases the complexity, inference time, and memory requirements of the models, it is relevant to examine the differences and advantages of multimodal compared to monomodal models. In this work, all available multimodal datasets for crowd counting are used to investigate the differences between monomodal and multimodal models. To do so, we designed a monomodal architecture that considers the current state of research on monomodal crowd counting. In addition, several multimodal architectures have been developed using different multimodal learning strategies. The key components of the monomodal architecture are also used in the multimodal architectures to be able to answer whether multimodal models perform better in crowd counting in general. Surprisingly, no general answer to this question can be derived from the existing datasets. We found that the existing datasets hold a bias toward thermal images. This was determined by analyzing the relationship between the brightness of optical images and crowd count as well as examining the annotations made for each dataset. Since answering this question is important for future real-world applications of crowd counting, this paper establishes criteria for a potential dataset suitable for answering whether multimodal models perform better in crowd counting in general. 

\end{abstract}

\subsection*{Keywords}
Crowd Counting, Multimodal Learning, RGB-T, Transformer

\vspace*{1.0\baselineskip}
}]


\section{Introduction}


\label{sec:intro}

One of the biggest challenges of crowd counting in real-world applications is dealing with varying lighting conditions. Since crowd counting can be very important for event security and crowd monitoring, good performance independent of lighting conditions is essential for real-world applications. Especially at night, lighting is often poor, resulting in less contrast and information in optical images and thus reducing the accuracy of prediction models. In this case, thermal images are more suitable because they do not rely on visible light. On the other hand, optical images can contain more information during the daytime compared to monochrome thermal images due to their color information. In addition, the environment may heat up during the day, resulting in lower contrast in thermal images, as human body temperature is almost constant. Overall, the use of both modalities seems to be symbiotic and to lead to better results compared to the use of a single modality. Using multiple modalities to train a model has led to state-of-the-art results in many cases. In particular, with the rise of transformers \cite{b1}, where inputs are transformed into homogeneous tokens, using multiple modalities such as text or images in a model has become easier. In the area of monomodal crowd counting, the use of transformers has not been fully explored. To our best knowledge, with the exception of one work \cite{b2}, previous research has focused only on convolutional networks. The use of transformers has tremendous potential, as previous work \cite{b3} \cite{b4} has often achieved better results when improving the extraction of multi-scale features. Although existing work \cite{b5} \cite{b6} concludes that the use of optical and thermal imagery leads to better crowd counting predictions, it is not yet fully understood how such models internally work and how they extract enriched features from both modalities.

Apart from the lack of understanding of how multimodal models work internally, it is not entirely understood whether the multimodal approach leads to better crowd counting results in general or only under certain conditions. Further research with potential influencing factors such as illumination, distance to the crowd, or number of people per image is needed to gain more certainty about whether multimodal crowd counting leads to better predictions in general. For this reason, in this paper we investigate the impact of using optical and thermal images simultaneously in crowd counting.

To investigate the impact of using optical and thermal images simultaneously in crowd counting, we designed a monomodal and several multimodal architectures consisting of the same key components. When we designed the monomodal model, we took into account the latest developments in the field of monomodal crowd counting. In addition, we have developed three multimodal models that incorporate different strategies of multimodal learning. To allow a comparison between the monomodal and the multimodal architectures, all key components of the monomodal architecture are also part of the multimodal architectures. The goal of this comparison is to find out whether multimodal models lead to better crowd counting results in general or only under certain conditions. Since this comparison led to interesting findings, we further analyzed all the datasets used to compare the models. To this end, we examined the relationship between the brightness of optical images and the number of individuals in the image. We also randomly selected a subset of each dataset and examined how individuals were labeled in the images from both modalities.

In examining the differences between the monomodal and the multimodal architectures, we found that existing datasets have a bias toward thermal images. This does not allow us to determine whether multimodal crowd counting leads to better results in general or only under certain conditions. For this reason, we have described criteria for a dataset suitable for investigating the research question.

\section{Related Work}
\textbf{Monomodal Crowd Counting:} Crowd counting has been studied for decades. While a few works have used thermal images for crowd counting, most works have used optical images to examine crowd counting. As in other areas, the use of deep learning models \cite{b18} \cite{b19} has led to more accurate predictions in crowd counting. In recent years, the use of a density map-based approach for crowd counting has become prevalent. Many recent works have addressed the question of how to deal with scale variations in images. In particular, techniques such as multi-column models \cite{b3} or dilated convolutions \cite{b4} have been used to extract multi-scale features from the image. Since such techniques aim to increase the receptive field of a network, it was no surprise that state-of-the-art results could be achieved by using a transformer encoder \cite{b1} to extract features \cite{b2}.

\textbf{Multimodal Crowd Counting:} Multimodal learning is becoming increasingly relevant in the field of crowd counting. So far, the use of optical and thermal images \cite{b5} \cite{b6} \cite{b7} \cite{b17} as well as the use of optical and depth images \cite{b8} \cite{b9} has been investigated. However, depth images provide only a limited depth range (0 $\sim$ 20 meters), making them unsuitable for many real-world crowd counting applications \cite{b5}. Also, when using depth images, there is still the problem that less information is available in poorly illuminated scenes. For this reason, we will focus on the use of optical and thermal images in this paper. While all work concludes that the additional use of thermal images leads to better predictions in crowd counting, it is not fully understood under what circumstances it is beneficial to complement optical images with thermal images to obtain better predictions. Previous work has focused primarily on constructing a novel model architecture that outperforms the state-of-the-art in multimodal crowd counting. While this approach proves the effectiveness of the models created, it does not allow us to fully understand how complementary information is extracted from both modalities.

\textbf{Multimodal Crowd Counting Datasets:} Similar to different multimodal models, two different datasets \cite{b5} \cite{b6} consisting of optical and thermal image pairs have been published in recent years. The dataset published by Peng et al. \cite{b6} was acquired with a drone and contains 3,600 image pairs. Furthermore, this dataset contains information about distance (scale of individuals), illumination and crowd count per image pair. The other dataset, which was published by Liu et al. \cite{b5}, contains 2,030 image pairs. The image pairs of this dataset were taken from a normal perspective. Information on the number of individuals and lighting is available for each image pair.


\section{Effectiveness of Multimodal Crowd Counting} To allow a comparison between monomodal and multimodal architectures, we first developed a monomodal architecture. This monomodal model takes into account recent advances in the field of monomodal crowd counting and its main components are reused in subsequent multimodal architectures to allow a fair comparison. Since the constructed monomodal architecture is heavily inspired by recent advances in monomodal crowd counting and does not incorporate any new strategies, we only used one monomodal model for comparison.

\subsection{Monomodal Architecure}
\begin{figure}[t]
    \centerline{\includegraphics[width=\linewidth]{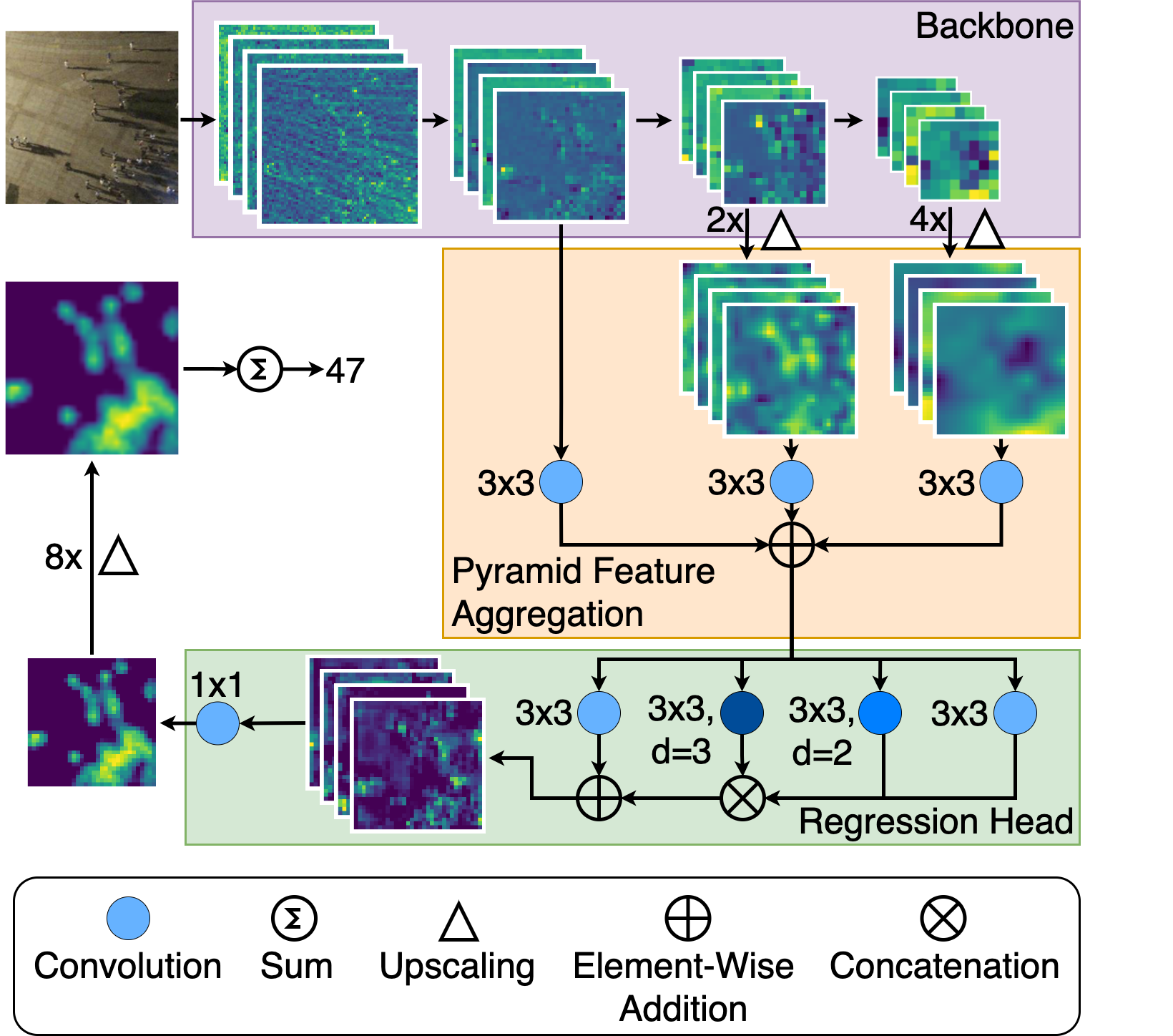}}
    \caption{The architecture of our monomodal model, which is inspired by the work of Tian et al. \cite{b2} as well as the implementation of the work realized in \cite{b10}. The input image is first transformed into tokens. From these tokens, features are extracted by a hierarchical transformer-based backbone. The hierarchical feature maps are then aggregated and finally used by the regression head to predict the crowd count. The parameter $d$ indicates the dilatation rate used.}
    \label{fig_monomodal_architecture}
\end{figure}

The monomodal architecture designed in this work is inspired by the work of Tian et al. \cite{b2} as well as the implementation of the work realized in \cite{b10}. The CCTrans model designed by Tian et al. \cite{b2} achieves state-of-the-art results on multiple monomodal crowd counting benchmarks \cite{b3} \cite{b11} \cite{b12}. Our monomodal architecture is shown in Fig. \ref{fig_monomodal_architecture}. Instead of Twins \cite{b13}, which was used by Tian et al. \cite{b2}, we used PVTv2 \cite{b14} as the transformer-based backbone in our architecture. By empirical analysis, we found that the PVTv2 architecture leads to better results for us. More specifically, for our monomodal architecture, we used the PVTv2 B0 variant, which allows shorter training time and requires less computational resources. However, this leads to slightly worse results compared to other PVTv2 variants with more parameters. This was acceptable to us, as our primary goal was not to construct a novel architecture with state-of-the-art results. Furthermore, we adopted the pyramid feature aggregation and regression head of Tian et al. \cite{b2}, but used the convolution kernel sizes used in \cite{b10}. Again, through empirical analysis, we found that these kernel sizes led to slightly better results for us.

\subsection{Multimodal Architectures}
After the monomodal architecture was designed, three different multimodal architectures were designed that incorporate different strategies of multimodal learning. As mentioned before, the key characteristics of the monomodal architecture are also incorporated in the three different multimodal architectures. The idea behind this is that when using the same weight initialization (prior) and the same model properties (which constrain the hypothesis space), better results can only be explained by more information provided by the additional modality (data). In this work, we chose to use early and late fusion as two simple multimodal strategies. These have also been used in previous work on multimodal crowd counting \cite{b5} \cite{b6}. In addition, we apply a more advanced deep fusion strategy using the Information Aggregation and Distribution Module (IADM) of Liu et al. \cite{b5} which has been shown to be effective for multimodal crowd counting.

\textbf{Early Fusion Model:} With the early fusion strategy, modalities are fused at the beginning of the model. For this purpose, the constructed monomodal model was adapted to support 6-channel inputs by changing the amount of filters in the first layer. Thus, the multimodal early fusion model has the same number of parameters as the monomodal model.

\textbf{Late Fusion Model:} In contrast to the early fusion strategy, the fusion of modalities takes place at the end of the model with the late fusion strategy. The idea here is that features of both modalities are first extracted individually. Thus, except for the final layer ($1 \times 1$ convolution), both modalities are investigated with the constructed monomodal model individually. Then, the extracted feature maps from both individual columns are concatenated. Based on the concatenated feature maps, a density map is then finally predicted by a $1 \times 1$ convolution. Hereby, the late fusion model requires around twice as many parameters as the monomodal model and the early fusion model.

\begin{figure}[t]
    \centerline{\includegraphics[width=\linewidth]{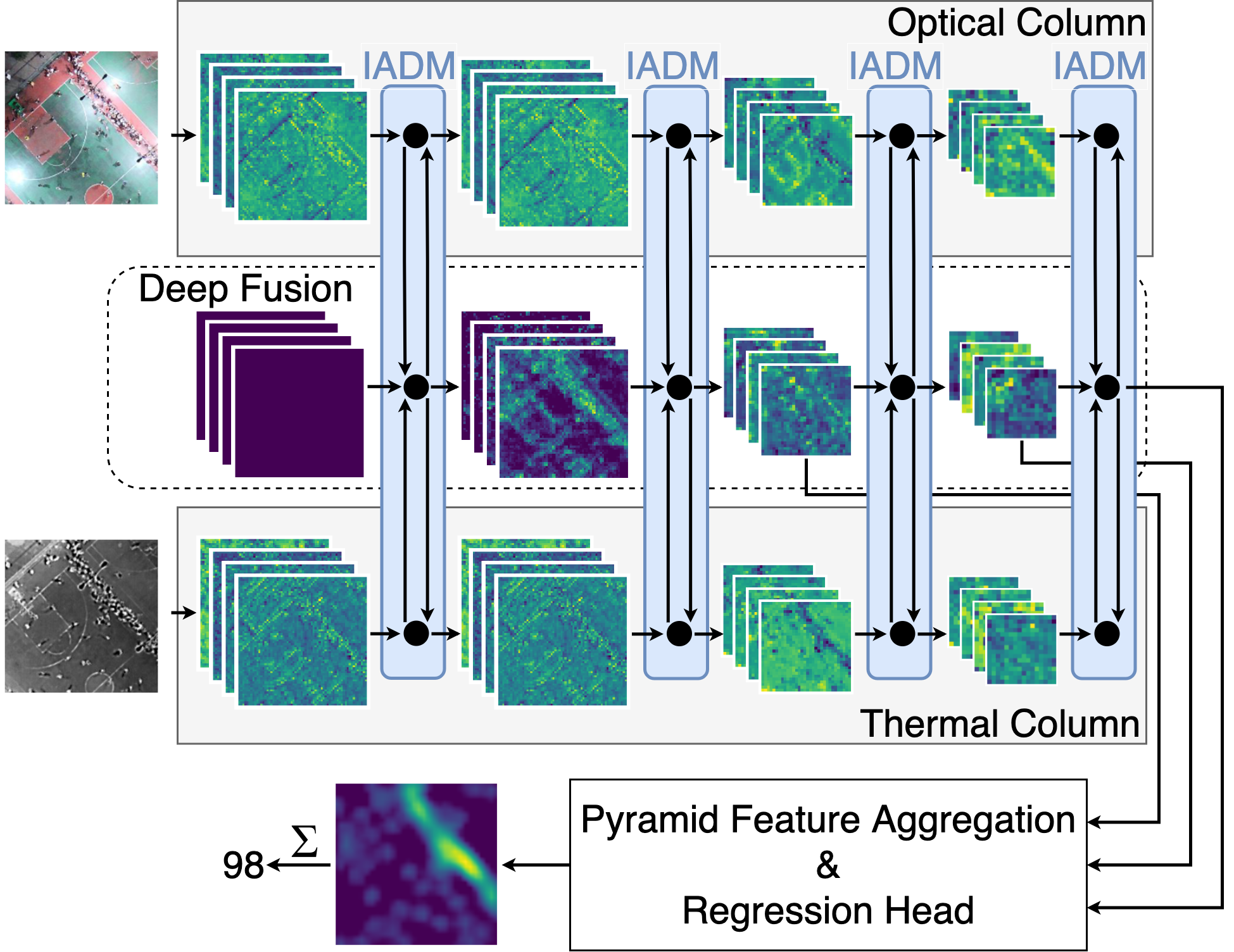}}
    \caption{The architecture of our deep fusion model. To extract complementary information and enable exchange between modality-specific and modality-shared columns, we use the IADM of Liu et al. \cite{b5}. In their work, it was shown that the use of the IADM is effective for multimodal data.}
    \label{fig_deep_fusion}
\end{figure}

\textbf{Deep Fusion Model:}
In contrast to the early fusion and late fusion architectures, the multimodal information exchange in the deep fusion architecture takes place during feature extraction. For this purpose, a third column is added to the architecture, which extracts the complementary information of both modalities. In particular, this is done by using the IADM of Liu et al. \cite{b5}. Through the IADM, information is exchanged between the modality-specific columns and the cross-modality column. However, this only takes place during feature extraction in the backbone, as shown in Fig. \ref{fig_deep_fusion}. Of all the models used in this work, this architecture requires the most parameters.

\subsection{Evaluation}
To evaluate the performance of the monomodal model and the three multimodal models, we used the mean absolute error (MAE) and the root mean squared error (RMSE). Both of these measures are widely used in crowd counting. The use of these measures allows comparison of our results with the results of other work. The mean absolute error and root mean squared error are defined as follows:

\begin{equation}
 MAE = \frac{1}{N} \sum_{i=1}^{N} | y_i - \hat{y}_i | \;,
\end{equation}

\begin{equation}
 RMSE = \sqrt{\frac{1}{N} \sum_{i=1}^{N} (y_i - \hat{y}_i)^2} \;,
\end{equation}
where $N$ is the number of image pairs, $y_i$ is the ground-truth number of individuals in image pair $i$, and $\hat{y}_i$ is the predicted number of individuals for image pair $i$.

\subsection{Training}
Overall, our training approach is heavily inspired by the training approach chosen by Tian et al. \cite{b2}. The B0 variant of the PVTv2 architecture was initialized with pre-trained weights in all experiments. We used random cropping with a cropping size of 256 for both dimensions and horizontal flipping with a probability of 50\% as augmentation strategies. In addition, AdamW \cite{b15} was used as the optimizer and a batch size of 8 was chosen for training. The learning rate was $1\mathrm{e}{-5}$ in all experiments, but was increasingly regulated by a weight decay of $1\mathrm{e}{-4}$. Bayesian loss \cite{b16} with a sigma value of 8 was used as a loss function. The models were trained for 60 epochs in all experiments.

\begin{table}[t]
\centering
\begin{tabular}{llrrr}
\hline
Modality & Architecture & MAE & RMSE \\
\hline
RGB  & Monomodal & 26.48 & 55.28\\
T  & Monomodal & 15.19 & 28.27\\
RGB-T  & Early Fusion & 14.92 & 25.86\\
RGB-T & Late Fusion & 13.83 & 25.16\\
RGB-T & Deep Fusion & 14.32 & 24.64\\
RGB-T & BL + IADM \cite{b5} & 15.61 & 28.18\\
RGB-T & TAFNet \cite{b7} & \textbf{12.38} & \textbf{22.45}\\
\hline
\end{tabular}
\caption{Performance of the different architectures on the RGBT-CC \cite{b5} dataset. The use of thermal images leads to dramatically better results compared to optical images. Moreover, the multimodal approach leads to better results than the monomodal approach.}
\label{table:results_rgbt_cc}
\vspace*{0.1cm}
\end{table}

\begin{table}[t]
\centering
\begin{tabular}{llrrr}
\hline
Modality & Architecture & MAE & RMSE\\
\hline
RGB  & Monomodal & 10.40 & 16.44\\
T  & Monomodal & \textbf{6.70} & \textbf{10.20}\\
RGB-T  & Early Fusion & 7.41 & 11.43\\
RGB-T & Late Fusion & 7.01 & 11.18\\
RGB-T & Deep Fusion & 7.20 & 11.45\\
RGB-T & MMCCN \cite{b6} & 7.27 & 11.45\\
RGB-T & MFCC \cite{b17} & 7.96 & 12.50\\
\hline
\end{tabular}
\caption{Performance of the different architectures on the Drone-RGBT \cite{b6} dataset. Surprisingly, using thermal images solely with the monomodal architecture led to the best result for the Drone-RGBT dataset. In contrast, using optical images solely with the monomodal architecture leads to considerably worse results.}
\label{table:results_drone_rgbt}
\end{table}

\subsection{Results}
The results for all constructed models on both datasets are shown in Tab. \ref{table:results_rgbt_cc} and Tab. \ref{table:results_drone_rgbt}. Three aspects in particular caught our attention, which we describe in more detail below.

\textbf{Discrepancy between optical and thermal images in both datasets.} One of the first things we noticed is that the monomodal model performs much better on thermal images than on optical images, as can be seen in Tab. \ref{table:results_rgbt_cc} and Tab. \ref{table:results_drone_rgbt}. This holds true for both datasets. Nevertheless, the discrepancy is larger for the RGBT-CC dataset than for the Drone-RGBT dataset. Since we used the exact same model and training approach, these results raise the question of whether thermal images are more suitable for crowd counting in general. Before investigating this question, we first wanted to gain a better understanding of both data sets. The investigation is described in more detail in Section \ref{dataset_investigation}.

\textbf{The monomodal model performs better than the multimodal models for the Drone-RGBT \cite{b6} dataset.} Contrary to our assumption that the multimodal approach of using optical and thermal images would lead to better crowd counting predictions, using thermal images solely led to the best result for the Drone-RGBT dataset. This result further affirmed our motivation to gain a better understanding of both datasets. To our best knowledge, we have achieved state-of-the-art results for the Drone-RGBT dataset using the monomodal architecture.

\textbf{IADM \cite{b5} seems to be less effective with transformer encoders.} Comparing the three multimodal models, the late fusion model achieves the best results on both datasets. The deep fusion model, although more complex and shown to be effective by Liu et al. \cite{b5}, performs worse in our study than the late fusion model. Since Liu et al. also compared the IADM to a late fusion model, the most obvious explanation for this is the use of a transformer encoder in our work. Liu et al. did not use a transformer encoder in their work. Nevertheless, a more detailed investigation beyond this work is needed to better understand why the IADM is less effective when used with transformer encoders.


\section{Analysis of Existing Multimodal Crowd Counting Datasets}
\label{dataset_investigation}
To understand more profoundly whether thermal images are better for crowd counting in general, or whether the characteristics of the datasets used lead to better results on thermal images, we used two different approaches.

\subsection{Relationship Between Brightness and Crowd Count} 
\begin{figure}[t]
    \centerline{\includegraphics[width=\linewidth]{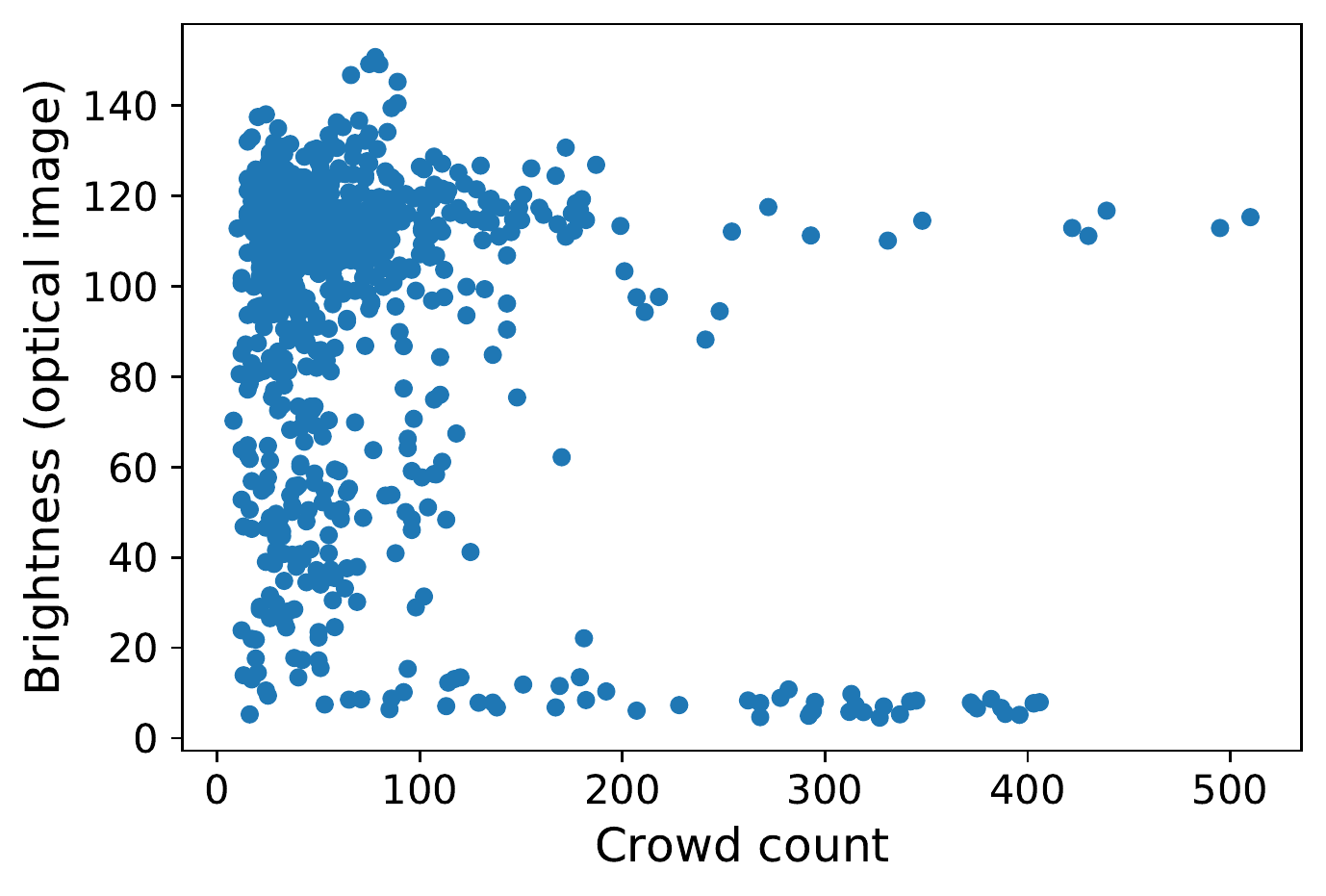}}
    \caption{Scatter plot showing the relationship between the brightness of optical images and crowd count in the RGBT-CC dataset. It can be seen that the RGBT-CC dataset is unbalanced in terms of brightness and crowd count. Many images with very low brightness have a high crowd count.}
    \label{fig_rgbt_cc_brightness_gt}
\end{figure}

\begin{figure}[t]
    \centerline{\includegraphics[width=\linewidth]{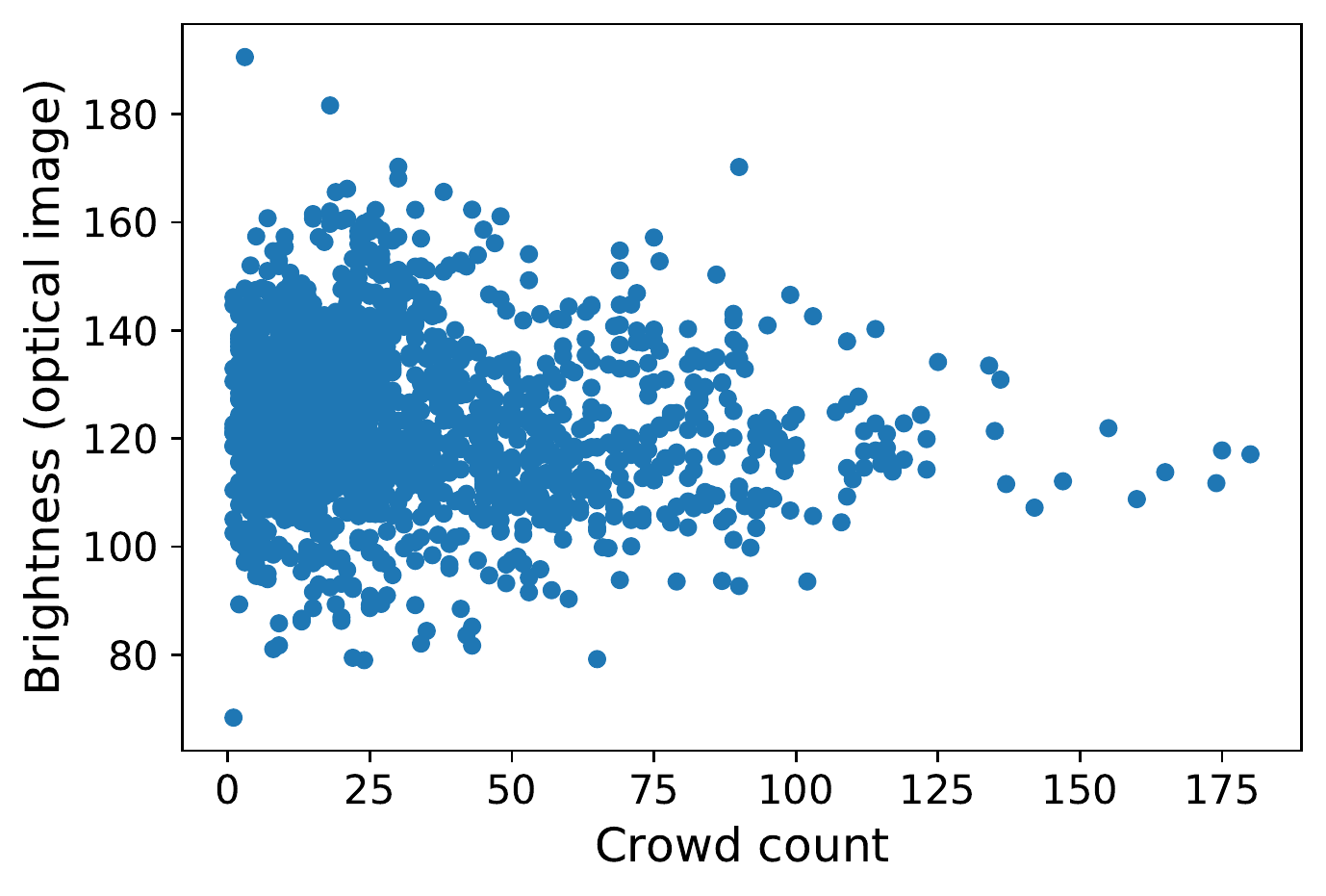}}
    \caption{Scatter plot showing the relationship between the brightness of optical images and crowd count in the Drone-RGBT dataset. The relationship between brightness and crowd count appears very uniform compared to the distribution of the RGBT-CC dataset.}
    \label{fig_drone_rgbt_brightness_gt}
\end{figure}

First, we investigated the relationship between the brightness of optical images and the number of individuals. We suspected that many optical images in both datasets were taken in poorly illuminated environments, which could be the reason for the discrepancy between thermal and optical images. This would also be in line with our main motivation to use multimodal data. Since the two metrics we used consider the counting error and are sensitive to outliers, we thought it is relevant to investigate the relationship between brightness and crowd count. To measure the brightness of an optical image, we used the following equation:

\begin{equation}
 Brightness = \frac{\sum_{i=1}^{W*H} R_i + G_i + B_i}{3*W*H}\;,
\end{equation}
where $W$ is the width and $H$ is the height of the optical image. $R_i$, $G_i$ and $B_i$ represent the three color values of pixel $i$. The relationship between brightness and crowd count for both datasets are shown in Fig. \ref{fig_rgbt_cc_brightness_gt} and Fig. \ref{fig_drone_rgbt_brightness_gt}.

\textbf{The RGBT-CC \cite{b5} dataset is unbalanced regarding brightness and crowd count.} The RGBT-CC dataset contains many images with very low brightness and high crowd count, as can be seen in Fig. \ref{fig_rgbt_cc_brightness_gt}. In comparison, the images in the Drone-RGBT dataset are much brighter on average and the overall distribution between brightness and number of individuals is much more balanced, as shown in Fig. \ref{fig_drone_rgbt_brightness_gt}. Since both metrics are sensitive to outliers and many optical images with very low brightness (low information) have a high crowd count in the RGBT-CC dataset, we assume that this explains the bigger discrepancy for the RGBT-CC dataset between optical and thermal images.

This finding has serious implications on our research question. Since this imbalance of the dataset likely affects all trained models and results in higher activations for thermal input, we believe that the research question cannot be thoroughly investigated with the RGBT-CC dataset. In particular, we assume that many optical images with low brightness (low optical information) and a high crowd count (high error) will cause the model to pay more attention to thermal images as the counting error is propagated back into the network during training. In this way, it is difficult to verify whether multimodal crowd counting leads to better results in general, since a certain condition (low brightness, high crowd count) has a great impact on the training of the model as well as the metrics. Nevertheless it is important to say that images with low brightness 
 and high crowd count are not a problem, but are important and desirable for training a robust crowd counting model. However, we are concerned about whether our research question can be fairly investigated due to an inherent correlation between the number of people and brightness in the RGBT-CC dataset.

\subsection{Annotation Sample Analysis}
Our second approach to better understand both datasets was to perform a sample analysis of how the annotations were made. Since both datasets contain two different modalities recorded with two different cameras, we wanted to understand if images of both modalities were synchronized and how perspective changes were handled (because the cameras were probably next to each other during the recording). We decided to perform the sample analysis of how the annotations were made since both datasets provide shared annotations for both modalities.
To this purpose, we randomly selected 10\% of the image pairs per dataset and visualized the annotations in the images of both modalities to verify how the individuals were labeled in each image.

\textbf{Only thermal images were used to label individuals in both datasets.} By randomly selecting 10\% of all image pairs per dataset and visualizing the annotations for both modalities, we found that both datasets used only the thermal image to label individuals. Examples of both datasets showing that only thermal images were used to label individuals are provided in the Appendix in Fig. \ref{rgbt_annotation_issues} and Fig. \ref{drone_rgbt_annotation_issues}.

\textbf{All image pairs of the Drone-RGBT dataset were taken at night.} We have seen that the optical images in the Drone-RGBT dataset are on average brighter than the optical images in the RGBT-CC dataset. However, by looking at the annotations for each image pair in the Drone-RGBT dataset, we perceived that all images were taken at night. To gain more confidence in this perception, we looked at all the optical images in the Drone-RGBT dataset. In this way, we found that all image pairs in the Drone-RGBT dataset were taken at night. Nevertheless, many images were taken in environments with much artificial light, therefore the optical images are on average brighter than those of the RGBT-CC dataset. The fact that all images were taken at night adds a new perspective to the results obtained with the Drone-RGBT dataset. This leads to the assumption that optical images do not provide additional information at night and that a monomodal approach with thermal images leads to better results. Further research beyond this paper is needed to validate this assumption. 

\textbf{For image pairs in the RGBT-CC dataset, individuals were sometimes visible in one modality but not the other.} Liu et al. \cite{b5} have already stated in their work that optical and thermal images in the RGBT-CC dataset are not strictly aligned because they were captured with different sensors. However, when examining the annotations, we found that not only were the image pairs not strictly aligned, but sometimes individuals were visible in one modality but not the other. Examples for this are provided in the Appendix in Fig. \ref{label_errors_example}.


\section{Criteria for a Multimodal Crowd Counting Dataset}
\label{criteria}
Because both datasets have some weaknesses that make it difficult to draw general conclusions about the effectiveness of multimodal crowd counting, we decided to set criteria for a suitable dataset. Overall, the image pairs should be taken evenly throughout the day. In this way, the variability of the two modalities gets extensively covered. Ideally, this would even take into account different seasons and climate zones. Also, the crowd count per image pair should be independent of when the image was taken. This allows for an equal influence of both modalities on the multimodal model during training, as no modality receives more attention due to a higher counting error. When labeling individuals, both modalities should be considered so that later models can learn to extract the information from both modalities and incorporate it into the prediction (even when one modality contains little information and the other contains much). Furthermore, the images for both modalities should be taken simultaneously. In this way, the images of both modalities are aligned as precisely as possible, which allows the use of the same annotations for both modalities.


\section{Is Multimodal Crowd Counting Better in General?}
The goal of this work was to find out if the simultaneous use of optical and thermal images leads to better predictions in crowd counting in general. We found that existing datasets have a bias toward thermal images, making it difficult to draw general conclusions about the effectiveness of multimodal crowd counting. 
The results on the Drone-RGBT dataset indicate that solely using thermal images at night results in better predictions than a multimodal approach. Since the RGBT-CC dataset contains both daytime and nighttime images, the better predictions with multimodal data seem to indicate that the multimodal approach leads to better results during the daytime. 
However, these assumptions are by no means proven, but could serve as hypotheses for future research. Furthermore, we encourage the creation of a multimodal dataset in order to be able to investigate such hypotheses. We have provided criteria for the creation of such a dataset in the previous Section \ref{criteria}. However, it remains an open question whether multimodal crowd counting (including technical challenges like perspective distortion and synchronization between modalities) is the perfect approach. It could also be the case that two monomodal models produce better results than one multimodal model.
For example, one monomodal model could be used with optical images during the day and another with thermal images at night.

\section{Conclusion}
In this work, we found that existing multimodal crowd counting datasets have a bias toward thermal images. For this reason, we outlined criteria for a balanced dataset. To our best knowledge, we also obtained state-of-the-art results on the multimodal Drone-RGBT dataset. Interestingly, for this we used solely thermal images and the monomodal model constructed in this work. Considering the results of this work, we encourage the creation of a multimodal dataset that meets the criteria outlined in this paper. In this way, we can understand more profoundly whether the simultaneous use of optical images and thermal images leads to better predictions in crowd counting in general.

%
%


\newpage
\onecolumn
\section*{Appendix}


\begin{figure}[h]
    \begin{minipage}{.49\textwidth}
      \includegraphics[width=\linewidth]{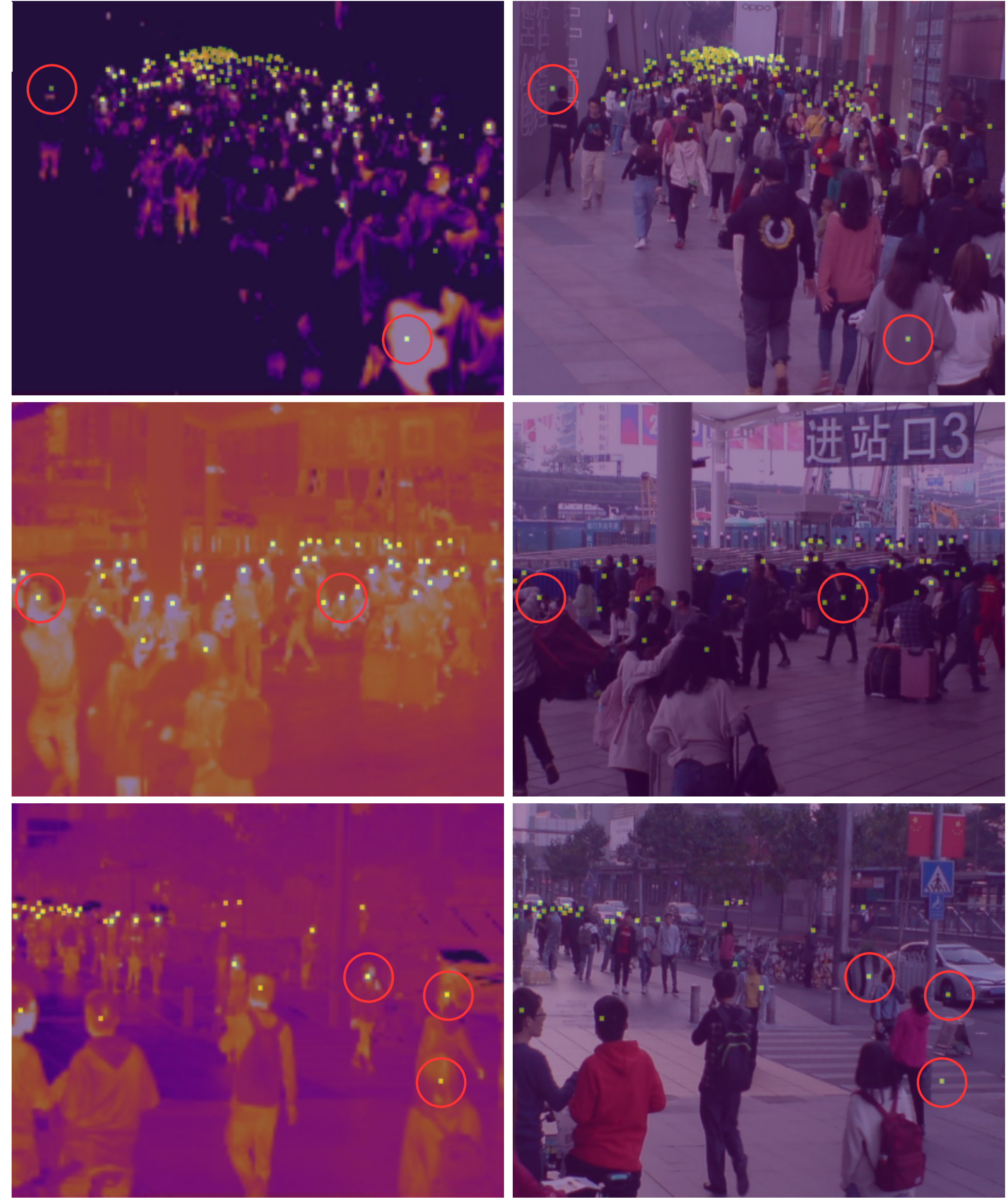}
      \caption{Illustration of the annotations (yellow squares) of the RGBT-CC \cite{b5} dataset for both modalities. It can be seen that the annotations were created based on the thermal images. For better understanding, examples of inaccuracies in the annotation of the optical images have been encircled. The thermal images are also encircled in the corresponding places, but the annotations are correct there.}
      \label{rgbt_annotation_issues}
      \vspace{.5cm}
    \end{minipage}%
    \hspace*{.02\textwidth}
    \begin{minipage}{.49\textwidth}
      \includegraphics[width=\linewidth]
      {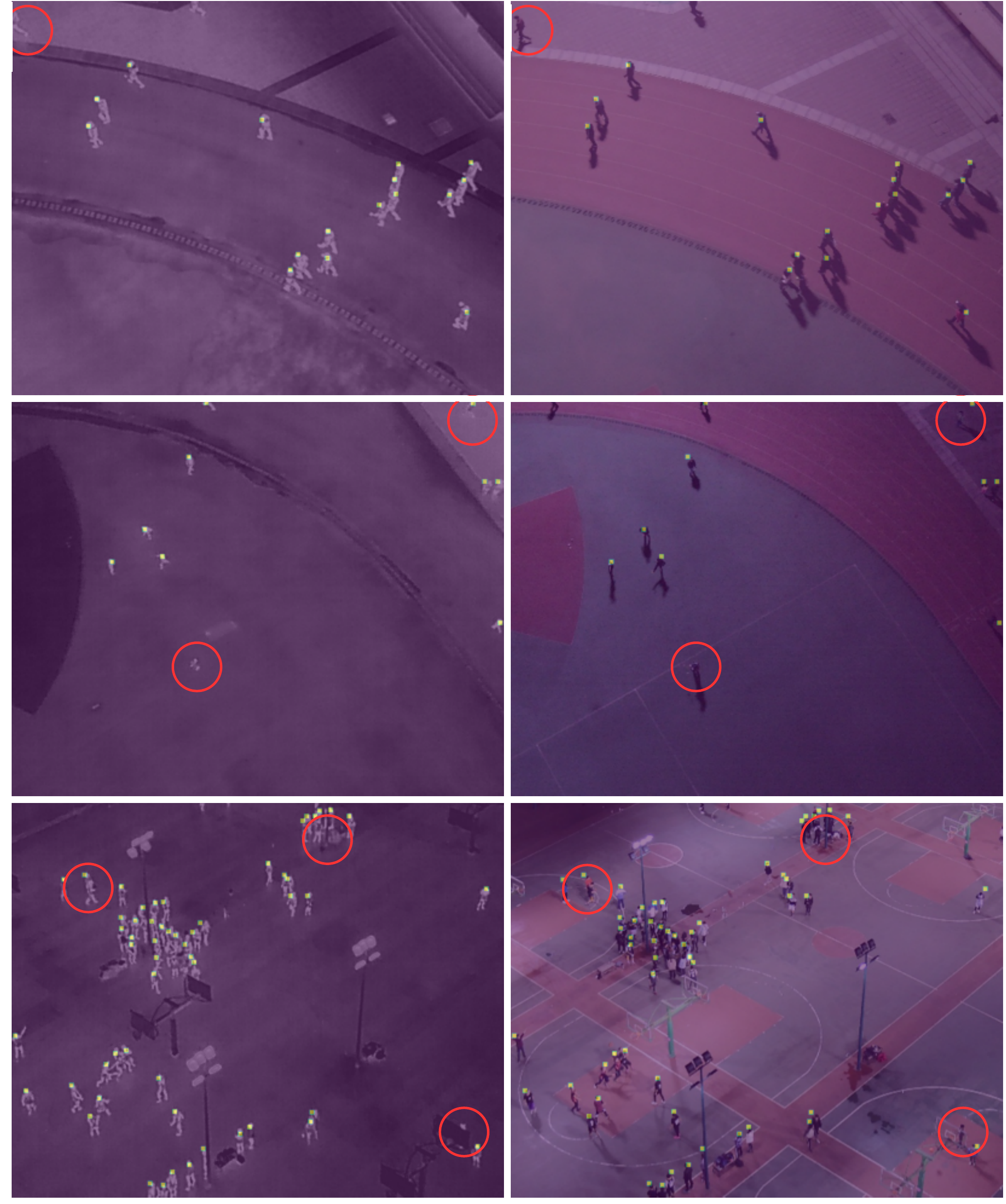}
      \caption{The annotations of the Drone-RGBT \cite{b6} dataset are shown in yellow squares. It can be seen that the annotations were made on the basis of the thermal images. For better understanding, individuals have been encircled who are easily recognizable in the optical image, but have not been annotated. In the thermal image, on the other hand, it can be seen that precisely these individuals are difficult to recognize, which is probably why they were not annotated.}
      \label{drone_rgbt_annotation_issues}
    \end{minipage}
\end{figure}


\begin{figure}[h]
    \centerline{\includegraphics[width=0.5\linewidth]{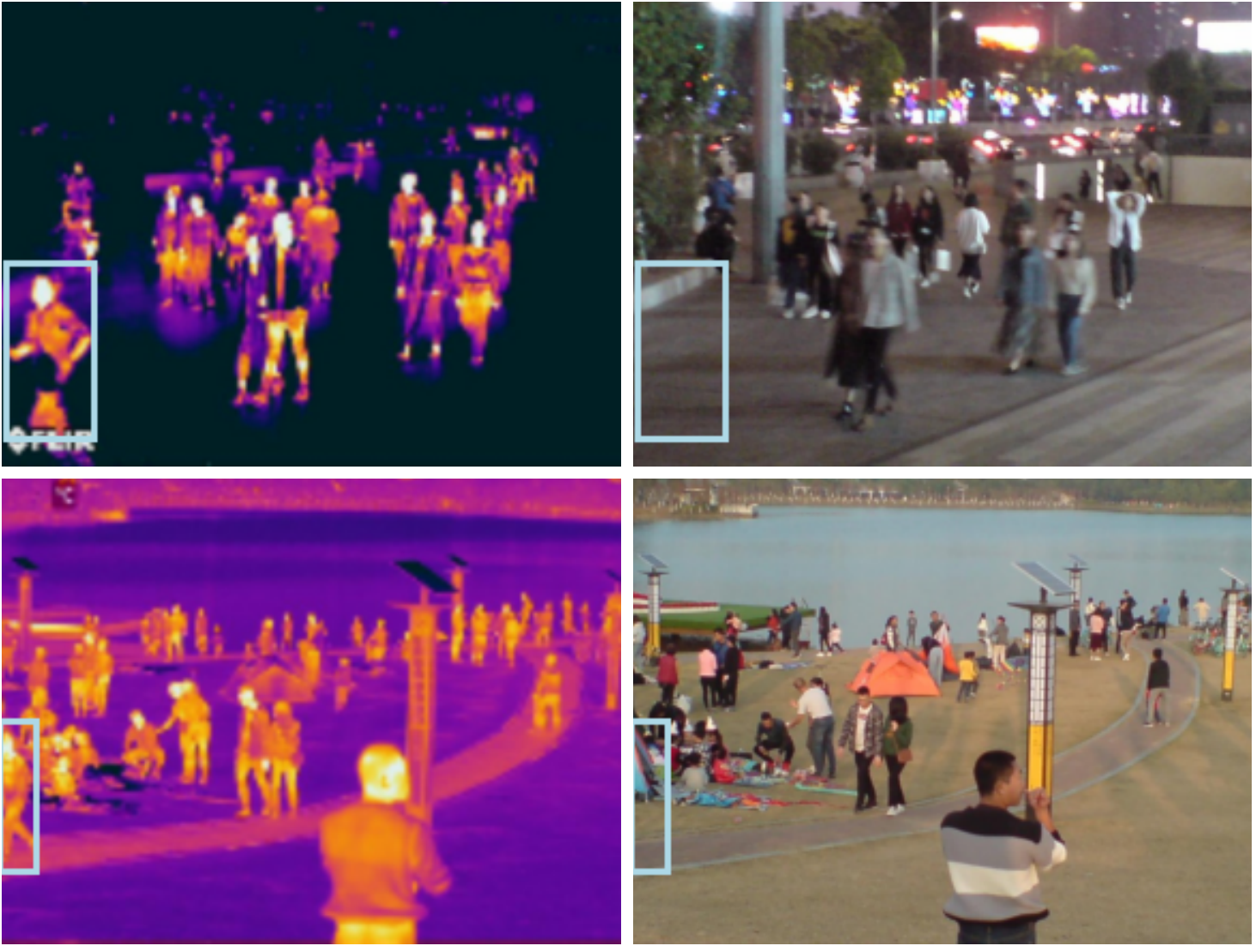}}
    \caption{Two examples from the RGBT-CC dataset where a person is visible in one modality but not the other. It can be seen that this is due to the time-delayed capture of the images from both modalities.}
    \label{label_errors_example}
\end{figure}

\end{document}